# A Note on Zipf's Law, Natural Languages, and Noncoding DNA Regions

Partha Niyogi and Robert, C. Berwick

*MIT Center for Biological and Computational Learning*

*Cambridge, MA 02139*

## Abstract

In *Phys. Rev. Letters*, **73:2, 5 Dec. 94**, Mantegna et al. conclude on the basis of Zipf rank frequency data that noncoding DNA sequence regions are more like natural languages than coding regions. We argue on the contrary that an empirical fit to Zipf's "law" *cannot* be used as a criterion for similarity to natural languages. Although DNA is a presumably an "organized system of signs" in Mandelbrot's (1961) sense, an observation of statistical features of the sort presented in the Mantegna et al. paper does not shed light on the similarity between DNA's "grammar" and natural language grammars, just as the observation of exact Zipf-like behavior cannot distinguish between the underlying processes of tossing an $M$ sided die or a finite-state branching process.

In *Phys. Review Letters*, **73:2, 5 Dec. 94**, Mantegna et al. "extend the Zipf approach to analyzing linguistic texts to the statistical study of DNA base pair sequences and find that the noncoding regions are more similar to natural languages than the coding sequences" (p. 3169). Specifically, the authors analyze coding/noncoding DNA sequences and conclude that noncoding regions show a more Zipf-like behavior than coding regions. Asserting that "A remarkable feature of languages is Zipf's law" (p. 3169), they further conclude that noncoding regions are more similar to natural languages than coding regions (p. 3170):

> The averages for each category support the observation that $\xi$ is consistently larger for the noncoding sequences, suggesting that the noncoding sequences bear more resemblance to a natural language than the coding sequences.

Their result has received popular notice in both *Science* (266, p. 1320, 25 Nov. 1994) and *Scientific American* (272(3), March, 1995).

In this note we would like to argue that the Mantegna et al. conclusion is rather far-fetched. Noncoding DNA sequences do *not* show much similarity to natural languages. Rather, as far as one can judge from the evidence of the Mantegna et al. paper, all one can say—if their statistical analysis is not in question, which it may well be—is that noncoding DNA sequences and natural languages combine discrete symbols to form strings that obey Zipf's law. But this is of course what we knew from the outset. In particular:

- Any number of random processes outputting discrete symbols can display Zipf-like behavior without bearing any resemblance to the special generative processes currently believed to govern sentence formation (word sequences) in natural languages. In this sense Zipf's law is not peculiar to natural languages at all, and therefore cannot be used as a strong test for whether DNA, or anything else for that matter, has something "in common with natural languages." Indeed, exactly this same point was made at length over 30 years ago by Mandelbrot (1961) in his familiar discussion of Zipf's law:

  > Further, because statistical and grammatical structures seem uncorrelated, in the first approximation, one might expect to encounter laws which are independent of the grammar of the language under consideration. Hence, from the viewpoint of significance (and also of the mathematical method) there would be an enormous difference between: *on the one hand*, the collection of data that are unlikely to exhibit any regularity other than the approximate stability of the relative frequencies, when different samples are compared [i.e., data leading to statistical laws like Zipf's law; our comments pn/rcb]; and, *on the other hand*, the study of laws that are valid for natural discourse [the discovery of such laws being the goal of linguistics pn/rcb] but not for other organized systems of signs. (p. 213)

As is also familiar and as we show by examples below, it is quite easy to generate Zipf-like distributions from very simple generative processes that are quite unlike natural languages, e.g., tossing an $M$-sided die and particular very simple finite-state branching processes.[1] In short, although DNA is a presumably an "organized system of signs" in Mandelbrot's sense, an observation of statistical features of the sort presented in the Mantegna et al. paper does not shed light on the similarity between DNA's "grammar" and natural language grammars, just as the observation of exact Zipf-like behavior cannot distinguish between the underlying processes of tossing an $M$ sided die or a finite-state branching process. An empirical fit to Zipf's law *cannot* be used as a criterion for similarity to natural languages.

- Zipf's law is given by $fr = C$ where $f$ is the frequency of any word, and $r$ is its rank, with words arranged from most frequent to least frequent. In other words $\ln(f) = K - \xi \ln(r)$, (with $\xi = 1$). The authors find that $\xi$ is 0.286 for coding regions, and 0.386 for noncoding regions, and 0.57 for natural languages. Without further statistical tests, it is not unreasonable to conclude that both coding and noncoding DNA sequences are more *alike* to each other than either is to natural languages, and that Zipf's law is *violated*. What is plainly required are the usual significance tests addressing precisely this question, e.g., the null hypothesis that coding $\xi$ is the same as natural language

[1] Indeed, as N. Chomsky points out (p.c.), if we take a collection of English sentences and define "words" by taking the strings starting with, say, "e" and ending with "e" then the resulting, more random collection of "words" shows a *better* fit to Zipf's "law"—precisely because there are no interfering effects from the more organized features of natural language words. On this view, the closer fit of noncoding sequences to a Zipf distribution actually means that noncoding DNA sequences are *more random* and *more unlike* natural languages than coding sequences—exactly the opposite conclusion that Mantegna maintain.

$\xi$. Since the variances are clearly available, the authors or others should be possible to carry the required tests on the original data.

- As a minor point, in fact the two measures used in the paper—Zipf behavior, and Shannon entropy—are exactly correlated. Therefore it is not surprising that given Zipf-like behavior for noncoding sequences, one would also observe that noncoding regions have lower entropy than coding regions. In effect, there is just one, not "two similar statistical properties" (p. 3172) that natural languages and noncoding sequences share (if they share it at all), namely, Zipf-like behavior (or lower entropy).

  For a finite number of "words," entropy is largest for a uniform distribution over word frequencies. The more skewed the word frequencies, the lower the entropy. For coding regions (with $\xi = 0.286$), the word frequencies fall off more slowly with rank than for noncoding regions ($\xi = 0.386$). Consequently, coding regions will have have higher entropy and lower redundancy than noncoding regions. Having carried out a Zipf analysis and obtained $\xi$, one does not need to compute a separate entropy test. Yet the authors do so (as they recognize implicitly in the caption of figure 3 of their paper).

Putting aside these and other possibly grave statistical fallacies, in the remainder of this note we exhibit two random processes, one an $M$-sided die, the other a finite-state grammar, that are very different from each other yet yield exact Zipf distributions. We then review some of the many properties of natural languages not shared by these two processes. Consequently, even if we accept the results of the Mantegna et al. paper, the inference from Zipf-behavior to a similarity with natural languages cannot be justified. As mentioned, these points have been discussed more than thirty years ago by Mandelbrot (1961), and we conclude with some historical remarks that underscore his results along with related, more recent work that has also examined Zipf-behavior in DNA sequences.

# I. ZIPF'S LAW AND RANDOM PROCESS: SOME EXAMPLES

## Zipf's Law and Random Processes

To begin, let us consider two very different, simple random processes that both generate Zipf distributions: an $M$-sided die and a finite-state grammar.

Let us first recall Zipf's "law" itself. Suppose there are $M$ "words" in a system. These words might be generated in various combinations according to some underlying process, giving rise to a corpus of sentences, or more generally, word sequences. Since there are only a $M$ words, each word would occur multiple times in a large (potentially infinite) corpus. One can then rank these words, from most frequent to least frequent. Let the frequency of the $i$th word be $f_i$. If $f_i$ is proportional to $\frac{1}{i}$, the generative process is said to obey Zipf's law.

### Example 1: An $M$-sided die.

Let the sequence of words be generated by throwing a biased $M$ sided die. In particular, let the die be such that the probability of the $i$th side appearing on top is given by:

$$Prob[i] = \frac{\frac{1}{i}}{\sum_{j=1}^{M} \frac{1}{j}}$$

Now consider the following process:

1. Toss the biased die.
2. If the die shows $j$, output word $w_j$.
3. Repeat 1.

Clearly, this process generates a sequence of words where the first word is twice as likely as the second, three times as likely as the third, and so on. The process thus follows Zipf's "law" exactly.

### Example 2: Finite-State Grammars

Next we consider a random process generating "sentences" in a completely different fashion from example 1, but still obeying Zipf's law. Rather than deal with the case of $M$ words directly, we provide some intuition in the form of an example where $M = 4$. Suppose there are four words: $w_1, w_2, w_3$, and $w_4$. Sentences (word sequences) are produced by combining words in some fashion according to a grammar. Let us assume that the generative process is as follows:

FIGURES

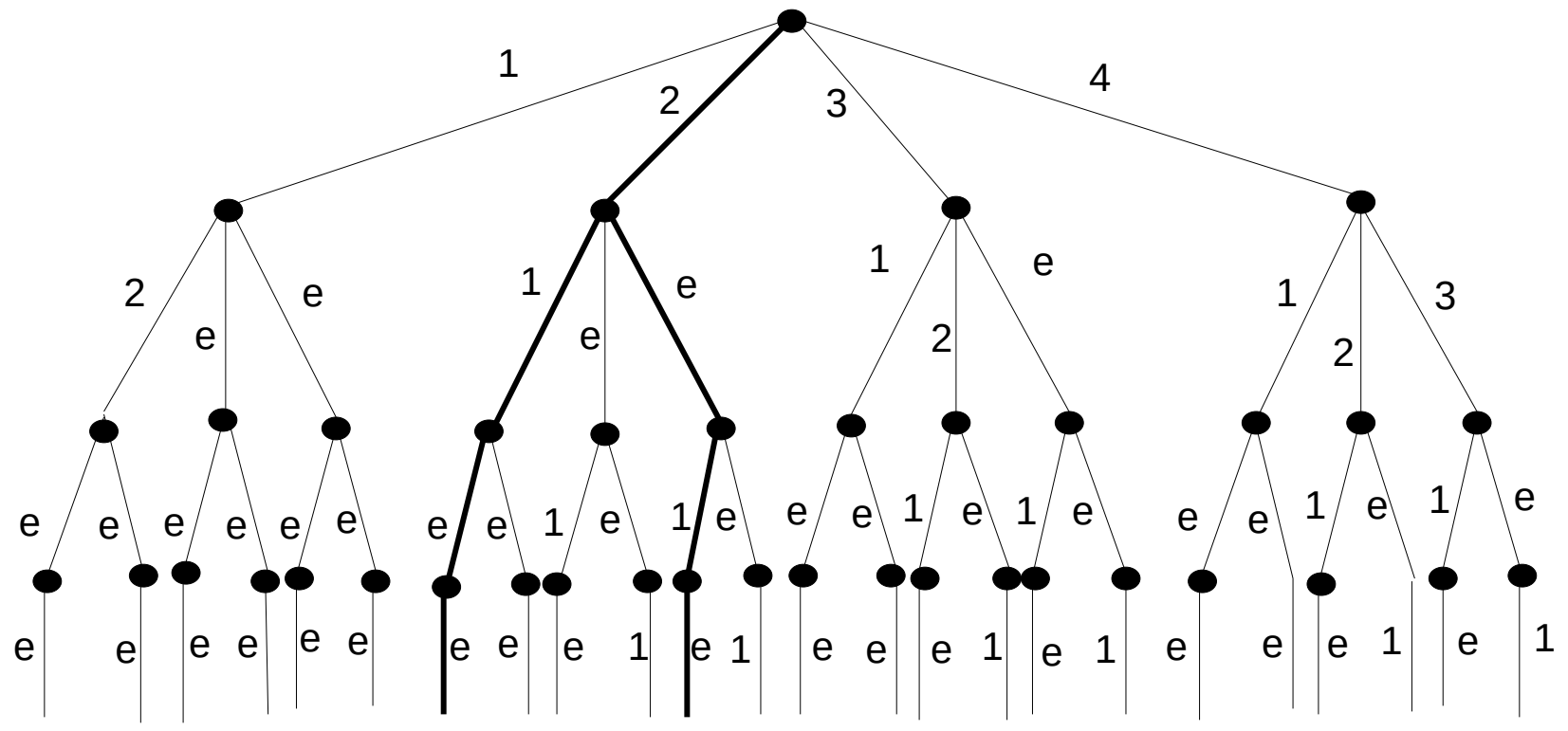

FIG. 1. A tree diagram representation of a finite state grammar.

1. Start at the root node of the annotated tree of fig. 1.

2. At each node, choose to go down any of the connected branches (leading to a daughter node) with equal probability. Output the word $w_i$ if the branch is associated with the number $i$. If the branch is associated with $e$, output nothing (empty string).

3. On reaching a leaf node, stop.

The reader will recognize that this is a finite-state grammar. Every path starting from the root node gives rise to a sentence. There are 4! different paths, corresponding to 4! different leaves, giving rise to 4! possible sentence types. Since the paths are all equally likely, each of these sentences occurs with equal likelihood.

However, due to the way in which the tree is constructed, many paths yield the same sentence. For example, the two paths highlighted in the figure yield the same sentence, $w_2 w_1$. The reader can check that such a grammar generates eight different sentences with the associated probabilities in table I.

## TABLES

| Sentence | $w_1$ | $w_1w_2$ | $w_2w_1$ | $w_3w_1$ | $w_3w_2w_1$ | $w_4w_1$ | $w_4w_2w_1$ | $w_4w_3w_1$ |
|---|---|---|---|---|---|---|---|---|
| Prob. | 1/6 | 1/12 | 1/4 | 1/6 | 1/12 | 1/12 | 1/12 | 1/12 |

TABLE I. Sentences generated by the finite state grammar of fig. 1, along with the probability with which they are generated.

If a corpus of sentences is generated with the probabilities shown in the table, then it can easily be shown that the word $w_1$ occurs twice as often as $w_2$, three times as often as $w_3$ and four times as often as $w_4$. In other words, if we plot word frequencies, then they would follow Zipf's law.[2]

In general, if there are $M$ words, then one could construct a similar tree. Such a tree would have $M!$ leaves, each leaf giving rise to a sentence. The branches could be numbered (as done in the case where $M = 4$) so that all the $M!$ different permutations of $M$ words can be generated. Now, as in the specific $M = 4$ case, we replace some of the numbers by $e$, equivalent to outputting an empty string for that branch. Let us now argue that this replacement can be carried out and yields a grammar that generates a Zipf distribution.

We first make the following observations to describe what $M$-tree looks like before any such replacements have been made. There are $M$ branches at level 1. Each of these branches bears a label from 1 to $M$, and no two branches bears the same label. There are $M(M-1)$ branches at level 2. There are an equal number of branches bearing each label from 1 to $M$. Consequently, $M-1$ of the branches at level 2, are labelled $i$ for every $i$ from 1 to $M$. Similarly, there $M(M-1)(M-2)$ branches at level 3, with $(M-1)(M-2)$ being labelled $i$ for every $i$ from 1 to $M$. As mentioned before, there are $M!$ different leaves, each giving rise to a different sentence (assuming no label were replaced by $e$). Each sentence is $M$ words long, a permutation of the $M$ words with no repeated word.

Next, consider how we replace the labels by empty strings $e$. Consider all the branches labelled $j$. Each time such a branch is traversed, the grammar outputs the word $w_j$. Suppose

---

[2]Note that the probability of occurrence of each word is inversely proportional to its rank. In a finite corpus, the frequency of occurrence need not be exactly equal to the probability. However, the convergence of frequencies to their underlying expectations make it more and more likely that frequency-rank behavior will follow Zipf's law as the number of sentences in the corpus increases, with convergence in the limit as the corpus size goes to infinity.

we chose to replace some of the $j$ labels by $e$, leaving only $a_1$ branches at level 1 still labelled, $a_2$ branches at level 2, and so on. We can then prove the following two theorems (given here without proof):

**Theorem .1** *Suppose $a_1$ branches at level 1 are still labelled and the remaining branches are labelled e. Similarly, suppose $a_2$ are labelled at level 2, $a_3$ labelled at level 3, and so on. Then a fraction $f$ of the total number of paths through the tree yields a sentence containing the word $w_j$, where $f$ is given by:*

$$\frac{a_1}{M} + \frac{a_2}{M(M-1)} + \frac{a_3}{M(M-1)(M-2)} + \ldots + \frac{a_M}{M!}$$

Clearly, $0 \leq a_1 \leq 1; 0 \leq a_2 \leq (M-1)$, and in general, $0 \leq a_i \leq \frac{(M-1)!}{(M-i)!}$. Given these constraints on the $a_i$'s, we can also prove the following:

**Theorem .2** *Any fraction that can be represented as $\frac{i}{M!}$ where $i$ is an integer between 0 and $M!$ can be obtained by an appropriate setting for the $a_i$'s under the constraints of Theorem 1.*

A consequence of these theorems taken together is that one can generate sentences in such a way that in a corpus the word $w_j$ can be made to occur in only a fraction $f = \frac{k}{M!}$ of the sentences. In particular, by choosing $k$ appropriately, we can make the $j$th word, $w_j$ occur with frequency $1/j$ in the text, thus following Zipf's law exactly.

## II. GENERAL REMARKS AND HISTORY

### A. Some Observations on the Structure of Natural Languages

It is well known that natural languages possess many other special properties that are not tested by the Zipf-law behavior. In particular, while finite-state grammars obey Zipf's law, it has long been known that they do not capture most of the striking properties of natural languages:

1. Finite-state grammars *by algebraic definition* cannot express hierarchical relationships, the acknowledged hallmark of natural languages. Recall that finite-state grammars are algebraically *associative* concatenative systems (see, e.g., Harrison, 1978); that is, if $\mathcal{L}$ is a finite-state grammar, then $\forall a, b, c \in \Sigma^*, a \cdot \ bc \in \mathcal{L} \ iff \ ab \cdot c \in \mathcal{L}$, where $\cdot$ is the concatenation operator. Such a system cannot even express the fact that one and the same linear string of words, such as "the deep blue sky" can have at least two structural (hierarchical) bracketings: (the (deep blue) sky) and (the deep (blue sky)). In other words, finite-state grammars can express only linear precedence relations, not hierarchical relations. (This demonstrates a failure of what Chomsky, 1956, called "strong generative capacity.")

2. Finite-state grammars, unlike natural language grammars, cannot generate arbitrarily deep center-embedded languages (see Chomsky 1956, 1986, and many other conventional sources).

3. Under the currently best working assumptions, natural language grammars contain very specific constraint statements with proprietary theoretical vocabularies unlikely to be duplicated in DNA "grammar," (e.g., one component, so-called "trace theory" is stated in terms of hierarchical structural sentence properties and noun phrases, both not shared by DNA, as far as it is known).[3]

### B. Previous work on Zipf's Law and on DNA word frequencies

Both Zipf's law and its application to DNA sequences have a long history. We mention only a few of the relevant points here. In the 1950s, as summarized in Mandelbrot (1961), both Mandelbrot, Simon (1955), and Miller and Newman (1958), among others, explored the nature of the word-frequency relationship embodied in Zipf's law. In particular, Mandelbrot showed how Markovian models of discourse (subsets of finite-state models) can give rise to Zipf-like behavior. Mandelbrot is careful to note the well-known inadequacy of such finite-state models to describe linguistic rules. For example, he writes (p. 191) "the 'finite-state' model appears as rather shocking because of the well known existence of some long-range influences in discourse, such as those studied by grammar". He advocates ways out of this difficulty while "acknowledging that the 'degree of validity' of the finite state model decreases as the 'wealth' of grammars increases." Mandelbrot also uses various information-theoretical arguments to suggest that Zipf's law is not peculiar to language, but extends to any coding scheme with a finite number of symbols—and therefore, can tell us relatively little about any coding scheme like DNA.

[3]We should point out that some researchers, e.g., Searls, 1993, maintain the contrary position and argue that natural language and DNA grammars share at least some generative processes. A discussion of this point is beyond the scope of this note.

As it turns out, there have also been many word-frequency analyses of DNA sequences. As Pevzner et al. (1989) point out, "Mathematical models of the generation of genetic texts appeared simultaneously with the first sequencing [of sic pn/rcb] DNA". Pevzner et al. (1989) actually address the key question of variance and significant differences explicitly, proposing formulae for the variance of number of word occurrences in texts, making it possible to assess the significance of deviations from expected statistical characteristics. One can therefore carry out the significance tests suggested earlier in this note.

## III. CONCLUSIONS

We have argued that an observation of Zipf-like behavior provides very little information about the nature of the underlying process generating such frequency data. This is simply because the underlying generative processes could be as diverse as $M$-sided dies, simple finite-state grammars, DNA sequences, and natural languages. Inferring that noncoding DNA sequence grammars are like natural language grammars solely on the basis of Zipf-behavior is at best premature, and indeed at worst is likely to be completely misleading and false.

**ACKNOWLEDGMENTS**

This note describes research done at the Center for Biological and Computational Learning and the Artificial Intelligence Laboratory of the Massachusetts Institute of Technology. Support for the Center is provided in part by a grant from the National Science Foundation under contract ASC–9217041. Robert C. Berwick was also supported by a Vinton-Hayes Fellowship.

This note has been improved by discussion with Morris Halle, Federico Girosi, Benoit Mandelbrot, and Noam Chomsky. All remaining errors are ours.